\documentclass[journal]{IEEEtran}
\ifCLASSINFOpdf
\else
\fi
\usepackage{times}
\usepackage{amsmath,amssymb,graphicx}
\usepackage{amsfonts}
\usepackage{url}
\usepackage{algorithm}
\usepackage{algorithmicx}
\usepackage{algpseudocode}
\usepackage{booktabs}
\usepackage{adjustbox}
\usepackage{threeparttable}
\usepackage{multirow}
\usepackage{subcaption}
\usepackage{caption} 
\usepackage{array}
\usepackage{color}
\usepackage{subcaption}
\usepackage[justification=centering]{caption}

\def\tblue#1{\textcolor[rgb]{0,0,1}{#1}} 
\def\tred#1{\textcolor[rgb]{1,0,0}{#1}}  

\graphicspath{{figure/}}

\usepackage{ragged2e}
\renewcommand{\raggedright}{\leftskip=0pt \rightskip=0pt plus 0cm}

\newcommand{\highlight}[0]{\color{blue}}
\begin{document}
%

\title{Fine-Grained Visual Classification via Simultaneously Learning of Multi-regional Multi-grained Features}

%
%
%

\author{Dongliang~Chang,
        Yixiao~Zheng,
        Zhanyu~Ma,
        Ruoyi~Du,
        and Kongming~Liang}
\markboth{Journal of \LaTeX\ Class Files,~Vol.~14, No.~8, August~2015}%
{Shell \MakeLowercase{\textit{et al.}}: Bare Demo of IEEEtran.cls for IEEE Journals}
%




\maketitle

\begin{abstract}
Fine-grained visual classification is a challenging task that recognizes the sub-classes belonging to the same meta-class. Large inter-class similarity and intra-class variance is the main challenge of this task. Most exiting methods try to solve this problem by  designing complex model structures to explore more minute and discriminative regions. 
In this paper, we argue that mining multi-regional multi-grained features is precisely the key to this task. 
Specifically, we introduce a new loss function, termed top-down spatial attention loss (TDSA-Loss), which contains a multi-stage channel constrained module and a top-down spatial attention module. The multi-stage channel constrained module aims to make the feature channels in different stages category-aligned. Meanwhile, the top-down spatial attention module uses the attention map generated by high-level aligned feature channels to make middle-level aligned feature channels to focus on particular regions. Finally, we can obtain multiple discriminative regions on high-level feature channels and obtain multiple more minute regions within these discriminative regions on middle-level feature channels. In summary, we obtain multi-regional multi-grained features.
Experimental results over four widely used fine-grained image classification datasets demonstrate the effectiveness of the proposed method. Ablative studies further show the superiority of two modules in the proposed method. \highlight{Codes are available at: https://github.com/dongliangchang/Top-Down-Spatial-Attention-Loss.}


\end{abstract}


%
\IEEEpeerreviewmaketitle

\section{Introduction}

\IEEEPARstart{F}{ine-grained} visual classification (FGVC) aims to recognize the sub-categories from one meta-class (\emph{e.g.}, bird species, car and aircraft models)~\cite{lin2015bilinear}. Compared to traditional image classification~\cite{luo2017convolutional,lei2019semi,li2020discriminative,cong2019going,lei2020deep}, fine-grained visual classification is much more challenging due to the larger inter-class similarity and intra-class variance~\cite{ding2019selective,fu2017look}.
One promising way to tackle this challenge is to find more subtle and discriminative regions from an input image.~\cite{wang2018learning, yang2018learning,luo2019cross,zheng2020iu}.

\begin{figure}[!t]
\begin{center}
  \includegraphics[width=0.9\linewidth]{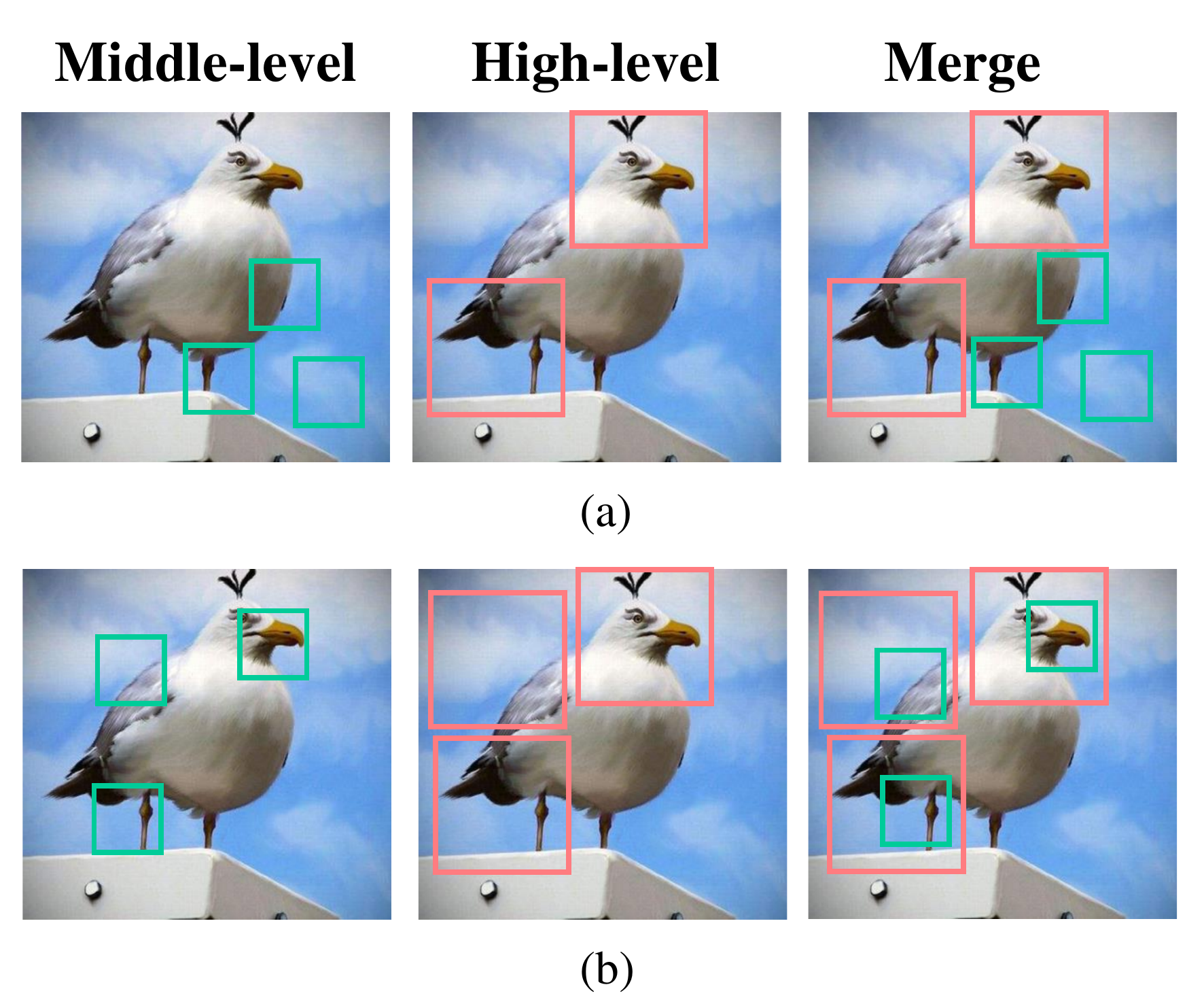}
\end{center}
  \caption{Illustrations of (a) one previous general method and (b) the proposed method for fine-grained visual classification.} 

\label{fig:insight}
\end{figure}

Some early works used hand-crafted bounding boxes or part annotations to assist the localization of discriminative and local regions~\cite{berg2013poof, xie2013hierarchical, branson2014bird, lei2016fast, li2019dual, ma2019fine}. However, expert knowledge is indispensable to the additional hand-crafted annotations and often error-prone~\cite{volkmer2005web}, making it expensive to implement and hinder practical deployment in real scenarios. Therefore, some researchers tried to learn part-level discriminate feature representations with only image-level category labels due to the shortcomings of additional hand-crafted annotations~\cite{dubey2018pairwise, lin2015bilinear, peng2017object,chang2020mc, wang2018learning,sun2018multi,luo2019cross}. 
Attention mechanism is often used by researchers to make the model focus on the most discriminative region~\cite{liu2020filtration,zheng2017learning,zhang2019learning,zheng2019looking,ji2020attention}. However, those methods ignore the other regions which are also helpful to learn the difference between sub-classes.
Other equally important methods noticed by researchers are to design task-specific loss functions to reinforce the learning ability of CNNs~\cite{chang2020mc,dubey2018maximum, dubey2018pairwise, gao2020channel, sun2020fine,zhuang2020learning}. The task-specific loss functions can implicitly make the model find the discriminative regions without increasing the model complexity. The exiting and future methods can easily integrated with the above loss functions.  However, those methods can only force the model to focus on one or several regions which are not enough for the FGVC task.

\begin{figure*}[!t]
\begin{center}
  \includegraphics[width=0.9\linewidth]{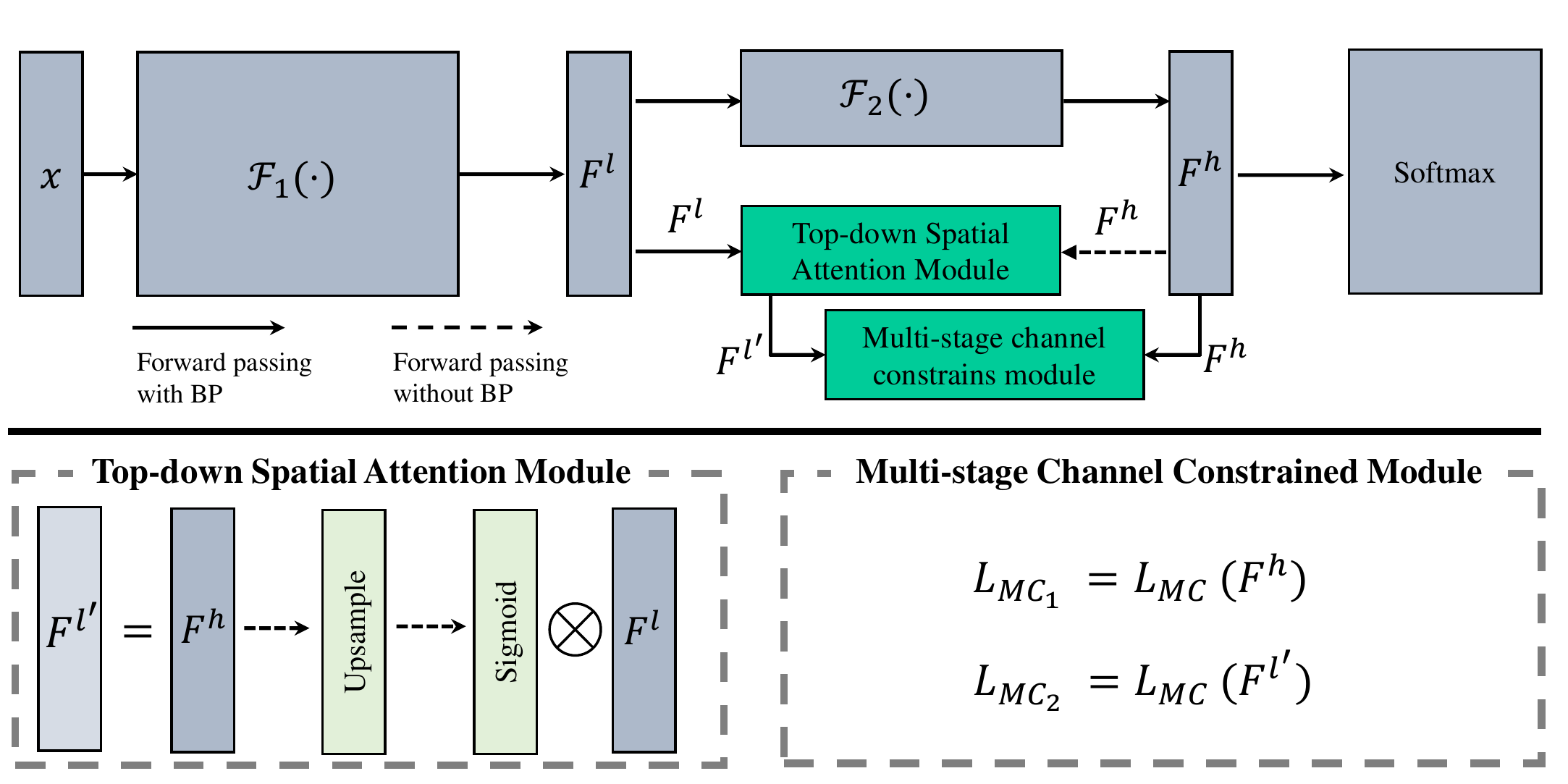}
\end{center}
  \caption{The overall framework with top-down spatial attention loss (TDSA-Loss). The proposed TDSA-Loss contains two modules: the top-down spatial attention module and the multi-stage channel constrained module.} 

\label{fig:overview}
\end{figure*}

In this paper, we also aim to localize discriminative local regions to deal with fine-grained image classification. Nevertheless, we argue that the multi-regional multi-grained features is the key to this task. To achieve this goal, the model should have two abilities. One is to locate multi discriminative regions, and the other is to mine multi-grained parts within different regions. The first step is relatively easy, since many existing algorithms can make the model find different and discriminative regions. However, the second step is challenging. Since the receptive field is fixed at each layer of the CNN, the multi-grained parts for each region is unable to obtain.

In general, the high-level features contain global concepts (\emph{e.g.}, bird's head, torso, or tail), the middle-level features contains local concepts (\emph{e.g.}, the beak or the eyes), and the low-level features discribe texture and shape.  Therefore, there has a directive way to locate multi-regional multi-grained features. Specifically, we can obtain the multi-global parts on the high-level features and get the multi-local regions on the middle-level features, comfortably realized by the existing methods. However, there 
exists an inconsistent problem: the obtained multi-local regions perhaps do not fall into the obtained multi-global parts, as shown in Figure~\ref{fig:insight}(a) .

To address the aforementioned problems, we attempt to connect the high-level and middle-level feature learning. By making the multi-global regions supervise the feature learning on the middle-level, we can force the model to extract the middle-level features within the regions proposed by the high-level. The multi-regional multi-grained features can be obtained following the procedure shown in Figure~\ref{fig:insight}(b). 
Specifically, we propose a new loss function, consists of a multi-stage channel constrained module and a top-down spatial attention module, termed top-down spatial attention loss (TDSA-Loss). The multi-stage channel constrained module is based on the mutual-channel loss~\cite{chang2020mc} and extends it to multi-stage, 
making the features in different stages category-aligned; the top-down spatial attention module uses the attention maps generated by the high-level aligned features to make the middle-level aligned features to focus on the small parts within the regions obtained by the high-level features. Finally, we can obtain multiple discriminative regions at high-level stages which are further used to obtain multiple local parts at middle-level stages -- multi-regional multi-grained features are all we need.

Extensive experiments are carried out on four commonly used fine-grained categorization datasets, CUB-$200$-$2011$~\cite{wah2011caltech}, FGVC-Aircraft~\cite{maji2013fine}, Stanford Cars~\cite{krause20133d}, and Flowers-$102$~\cite{nilsback2008automated}. The results show that our method can outperform the current state-of-the-art by a significant margin. Ablative studies are further conducted to verify the effectiveness of each of the proposed loss components and hyper-parameters.



\begin{figure*}[!t]
  \centering
   \begin{subfigure}[t]{2.5in}
    \centering
    \includegraphics[width=2.5in]{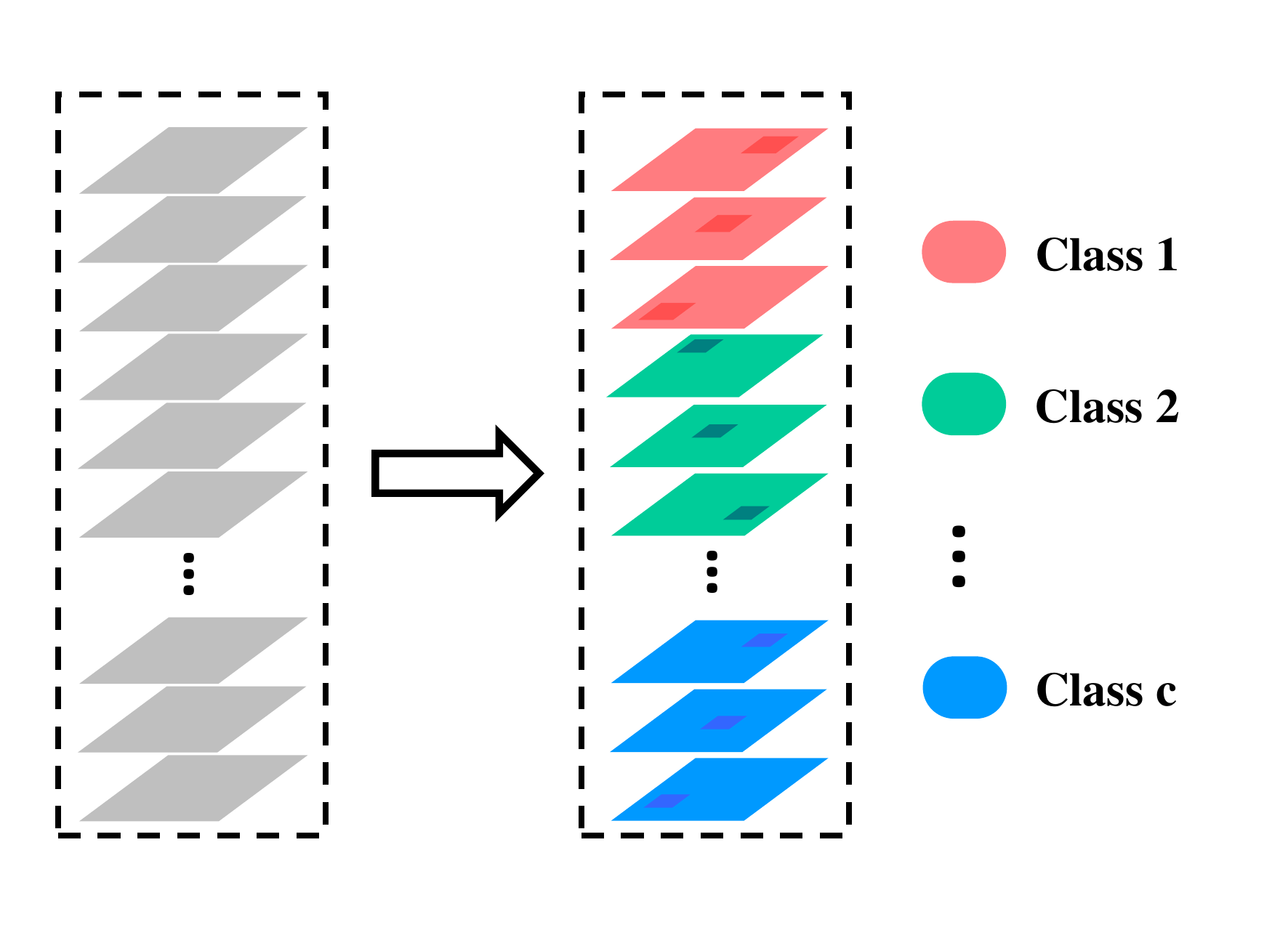}
    \caption{}
 \end{subfigure}
  \begin{subfigure}[t]{3.5in}
    \centering
    \includegraphics[width=3.5in]{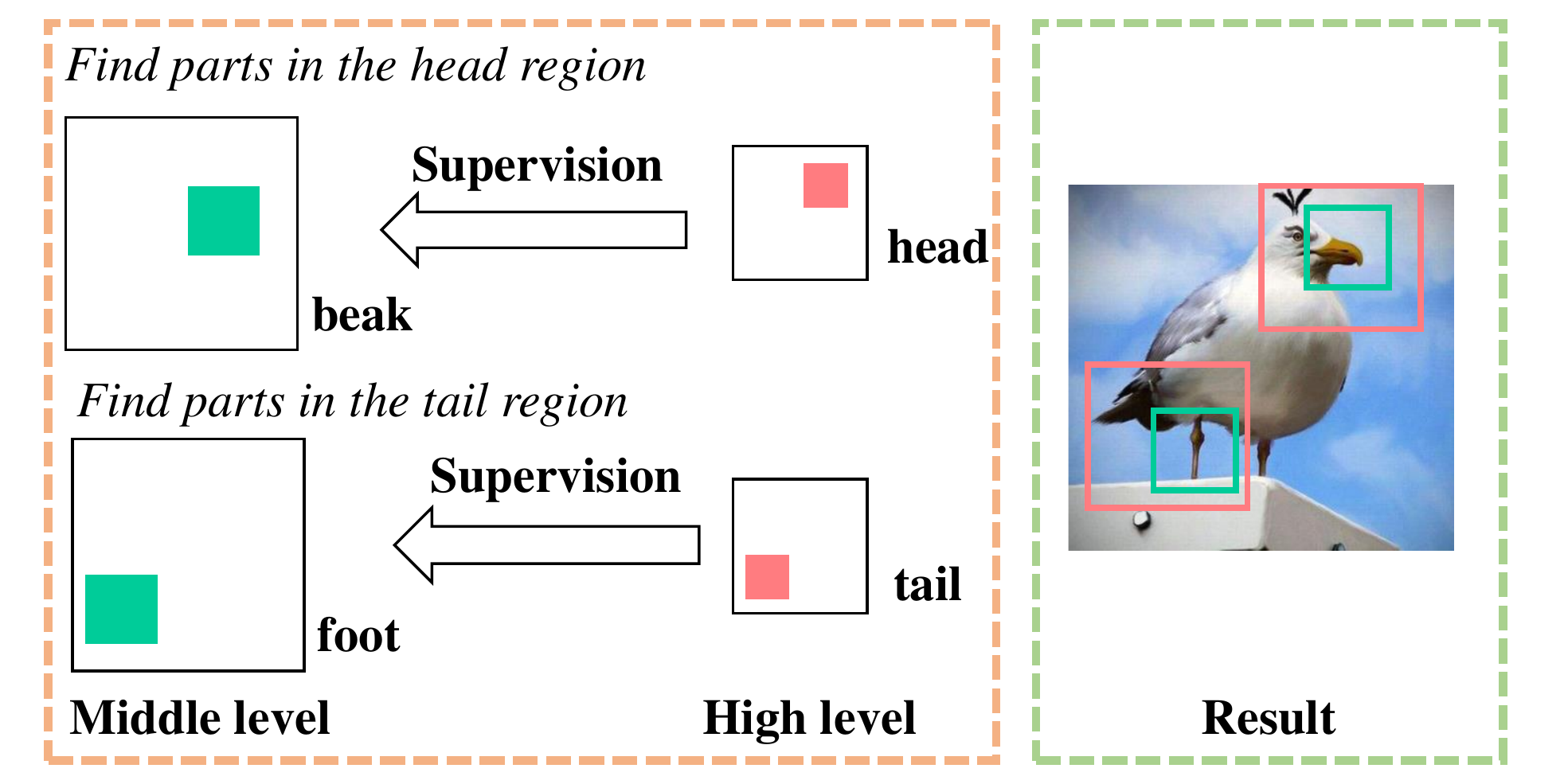}
    \caption{}
 \end{subfigure}

  \caption{(a) Comparison of feature maps before(left) and after (right) applying MC-Loss~\cite{chang2020mc}, where features become class aligned, and each is attending to different discriminate parts. (b) Illustration of the role of the proposed top-down spatial attention loss.}
  \label{fig:architecture}
\end{figure*}

\section{Related Work}
%
%

Some of the early works used hand-crafted bounding boxes or part annotations as additional information to assist the localization of discriminative and local regions~\cite{berg2013poof, xie2013hierarchical, branson2014bird, lei2016fast, li2019dual, ma2019fine}. However, expert knowledge is indispensable to the hand-crafted additional annotations, which makes it expensive to implement and hinders practical deployment in the real scenarios. Due to the shortcomings of hand-crafted additional annotations, some researchers tried to learn part-level discriminate feature representations with only image-level category labelss~\cite{dubey2018pairwise, lin2015bilinear, peng2017object,chang2020mc, wang2018learning,sun2018multi,luo2019cross,du2020fine}. Lin~\emph{et al.}~\cite{lin2015bilinear} proposed the bilinear model, whose outputs are multiplied using outer product at each location of the image and pooled to obtain an image descriptor. This architecture can model local pairwise feature interactions in a translationally invariant manner and allows end-to-end training with image labels only. Wang~\emph{et al.}~\cite{wang2018learning} designed a novel asymmetric multi-stream architecture and train a bank of convolutional filters to capture class-specific discriminative patches without extra part or bounding box annotations. Du~\emph{et al.}~\cite{du2020fine} proposed the Progressive Multi-Granularity Training of Jigsaw Patches, including a novel progressive training strategy that adds new layers in each training step to exploit information based on the smaller granularity information found at the last step and the previous stage and a simple jigsaw puzzle generator to form images contain information of different granularity levels.

Among them, one of the popular way applied by researchers is the well-designed attention mechanism~\cite{liu2020filtration,zheng2017learning,zhang2019learning,zheng2019looking,ji2020attention}. Fu~\emph{et al.}~\cite{fu2017look} proposed a novel recurrent attention convolutional neural network (RA-CNN), which recursively learns discriminative region attention and region-based feature representation at multiple scales. Zheng~\emph{et al.}~\cite{zheng2017learning} introduced a novel part learning approach by a multi-attention convolutional neural network (MA-CNN), which can generate more discriminative parts from features and learn better fine-grained features from parts in a mutual reinforced way. Sun~\emph{et al.}~\cite{sun2018multi} designed a novel attention-based convolutional neural network (CNN) which regulates multiple object parts among different input images by pulling same-attention same-class features closer and pushing different-attention or different-class features away. Ding~\emph{et al.}~\cite{ding2019selective} proposed the Sparse Sampling Networks (S3Ns), which collects peaks from class response maps to estimate informative receptive fields and learns a set of sparse attention for capturing fine-detailed visual evidence as well as preserving context.

Another equally important method noticed by researchers is to design task-specific loss functions to reinforce the learning ability of CNNs~\cite{chang2020mc,dubey2018maximum, dubey2018pairwise, gao2020channel, sun2020fine,zhuang2020learning}. Dubey~\emph{et al.}~\cite{dubey2018pairwise} proposed Pairwise Confusion (PC) and construct a Siamese neural network trained with a novel loss function that attempts to introduce confusion in output logit activations and prevents the network from overfitting to sample-specific artifacts. Li~\emph{et al.}~\cite{li2019dual} added a regularization term to the cross-entropy loss and propose a new loss function, Dual Cross-Entropy Loss. The regularization term places a constraint on the probability that a data point is assigned to a class other than its ground-truth class, which can alleviate the vanishing of the gradient when the value of the cross-entropy loss is close to zero. Chang~\emph{et al.}~\cite{chang2020mc} introduced the mutual-channel loss (MC-Loss), which consists of a discriminality component and a diversity component. The discriminality component forces all features  belonging to the same class to be discriminative and the diversity component additionally constraints features so that they become mutually exclusive across the spatial dimension.

Unlike the aforementioned methods, the proposed top-down spatial attention loss is a combination of well-designed attention mechanism and task-specific loss functions. 
We extend the mutual-channel loss~\cite{chang2020mc} to multi-stage, which can make the features in different stage category-aligned.
With the proposed top-down spatial attention module, middle-level convolutional filters can dig finer discriminative features under the supervison of high-level convolutional filters. In the meanwhile, strengthened middle-level features are beneficial to high-level features. In the proposed framework, more discriminative high-level and finer middle-level features can be collected mutually.

\section{The Proposed Approach}\label{section:Approach}
In the proposed framework shown in Figure~\ref{fig:overview}, we can see that the proposed loss function contain two modules: the top-down spatial attention module and the multi-stage channel constrained module. The multi-stage channel constrained module can be viewed as an extension of mutual-channel loss in~\cite{chang2020mc}. 
From the Figure~\ref{fig:architecture}.a, we can see that: with the mutual-channel loss, a model can effectively focus on different discriminative regions without any bounding-box or part annotations, and the features will become category-aligned. 
Especially, we impose spatial attention supervision obtained from a high-level convolutional layer on middle-level features and navigate the middle-level convolutional layer to search discriminative features in certain semantic regions, as shown in~\ref{fig:architecture}.b. In order to ensure sufficient discriminative information, we apply mutual-channel loss on both these different convolutional layers, \emph{e.g.}, $10^{th}$ convolutional layer conv4\_3 and $13^{th}$ convolutional layer conv5\_3 in a VGG-16 network~\cite{simonyan2014very}. The algorithm of the proposed method is summarized in Algorithm~\ref{alg:method}.

In the rest of Section~\ref{section:Approach}, we first review mutual-channel loss in Section~\ref{section:Mutual-channel Loss} and then introduce the proposed top-down spatial attention loss in Section~\ref{section:Cross-layer}. 

\begin{algorithm}[t]
  \caption{Top-Down Spatial Attention Loss $L_{TDSA}$}
  \hspace*{0.0in} {\bf Input:}
  training set $D = \{x_i, y_i\}_{i=1}^N$, having $N$ labeled examples \\
  \hspace*{0.0in} {\bf Initialize:} 
  operations between input and middle-level features $\mathcal{F}_1(\cdot)$, operations between middle-level features and high-level features $\mathcal{F}_2(\cdot)$, classifier $C$, weight coefficient $\mu$, $\lambda$, and the max iterations $max\_iter$\\
  \hspace*{0.0in} {\bf Output:} 
  loss function
  \begin{algorithmic}[1]
    \For{1 : $max\_iter$} 
    \State Randomly sample a batch data $(x, y)$ from $D$
    \State $F^l = \mathcal{F}_1(x)$
    \State $F^h = \mathcal{F}_2(F^l)$
    \State \# Top-down Spatial Attention Module
    \State ${F^l}^{\prime} = F^l \otimes Sigmoid(Upsample(F^h))$
    \State \# Multi-stage Channel Constrained Module
    \State $L_{TDSA} = L_{{MC}_1}(F^h,y) +  L_{{MC}_2}({F^l}^{\prime},y)$
    \State $L_{CE} = Cross\_Entropy\_Loss(C(F^h),y)$
    \State $L_{total} = L_{CE} + {\mu} \times L_{TDSA}$
    \State \Return  $L_{total}$
    \EndFor
  \end{algorithmic}
\label{alg:method}
\end{algorithm}

\subsection{Mutual-channel Loss}\label{section:Mutual-channel Loss}
In the fine-grained visual classification task,
the training set can be defined as $D = \{x_i, y_i\}_{i=1}^N$, where $N$ is the number of samples. Besides, the number of categories can be defined as $S$. Aiming at leading the model to focus on different discriminative regions, Chang~\emph{{et al.}}~\cite{chang2020mc} proposed the mutual-channel loss that consists two key components, \emph{i.e.}, discriminality component $L_{dis}$ and diversity component $L_{div}$. The mutual-channel loss $L_{MC}$ is added to the cross entropy loss $L_{CE}$ with the weight of $\mu$ in the training step and is also formulated as a weighted summation of its two components:
\begin{equation}
L_{MC} = L_{dis}(F, y) - \lambda \times L_{div}(F),
\end{equation}
\begin{equation}
L_{total_{MC}} = L_{CE} + \mu \times L_{MC},
\end{equation}
where $L_{total_{MC}}$ is the total loss function of the whole network. $F \in \mathbb{R}^{C \times H \times W}$ represents the feature maps outputed by a convolutional layer. $\lambda$ is a weight coefficient.

\subsubsection{The Discriminality Component}

The discriminality component is designed to enforce each channel of feature maps to be class-aligned and discriminative enough. According to the number of categories, features are divided into $S$ groups, \emph{i.e.}, $F = [F^1, F^2, ..., F^i, ...,F^S]$, $F^i \in \mathbb{R}^{\vartheta \times H \times W}$. $\vartheta$ denotes the number of features assigned to $i^{th}$ category. Consisting of channel-wise attention (CWA), cross-channel max pooling (CCMP), global average pooling (GAP), \emph{etc}, the discriminality component $L_{dis}$ is formulated as
\begin{equation}
g(F^i) = \underbrace{\frac{1}{H \times W} \sum_{j=1}^{H} \sum_{k=1}^{W}}_{\text{GAP}}
\Big(
\underbrace{\max_{m=1,2,...,\xi}\vphantom{\sum_{j=1}^{W}}}_{\text{CCMP}}
\big(
\underbrace{M_i \odot F_{m,j,k}^{i}\vphantom{\sum_{j=1}^{W}}}_{\text{CWA}}
\big)
\Big),
\end{equation}
\begin{equation}
L_{dis} = L_{CE}(Softmax(g(F^1), g(F^2), ..., g(F^S)), y),
\end{equation}
where $M_i = diag(Mask_i)$. $M_i \in \mathbb{R}^{\xi}$ is a 0-1 mask with randomly $\big\lfloor \frac{\xi}{2} \big\rfloor$ zero(s). The $\big\lceil \frac{\xi}{2} \big\rceil$ ones and operation $diag(\cdot)$ puts a vector on the principle diagonal of a diagonal matrix. $\odot$ denotes the matrix-vector multiplication.

\subsubsection{The Diversity Component}

The diversity component is designed to drive the features in a group $F^i$ to become different from each other and prevent all the features from focusing on the same discriminative region. By diversifying the features in each group, the diversity component helps to discover different discriminative regions with respect to every class in an image. Introducing the CCMP to measure the degree of intersection within each group, the diversity component can be formulated as
\begin{equation}
h(F^i) = \sum_{j=1}^{H} \sum_{k=1}^{W}
\underbrace{\max_{m=1,2,...,\xi}\vphantom{\sum_{j=1}^{W}}}_{\text{CCMP}}
\Bigg(
\underbrace{\frac{e^{F^i_{m,j,k}}}{\sum_{j^{\prime}=1}^{H} \sum_{k^{\prime}=1}^{W}e^{F_{m,j^{\prime},k^{\prime}}^i}}\vphantom{\sum_{j=1}^{W}}}_{\text{Softmax}}
\Bigg),
\end{equation}
\begin{equation}
L_{div}(F) = \frac{1}{S} \sum_{i=1}^{S} h(F^i).
\end{equation}

\begin{figure*}[!t]
    \begin{center}
   \includegraphics[width=0.95\linewidth]{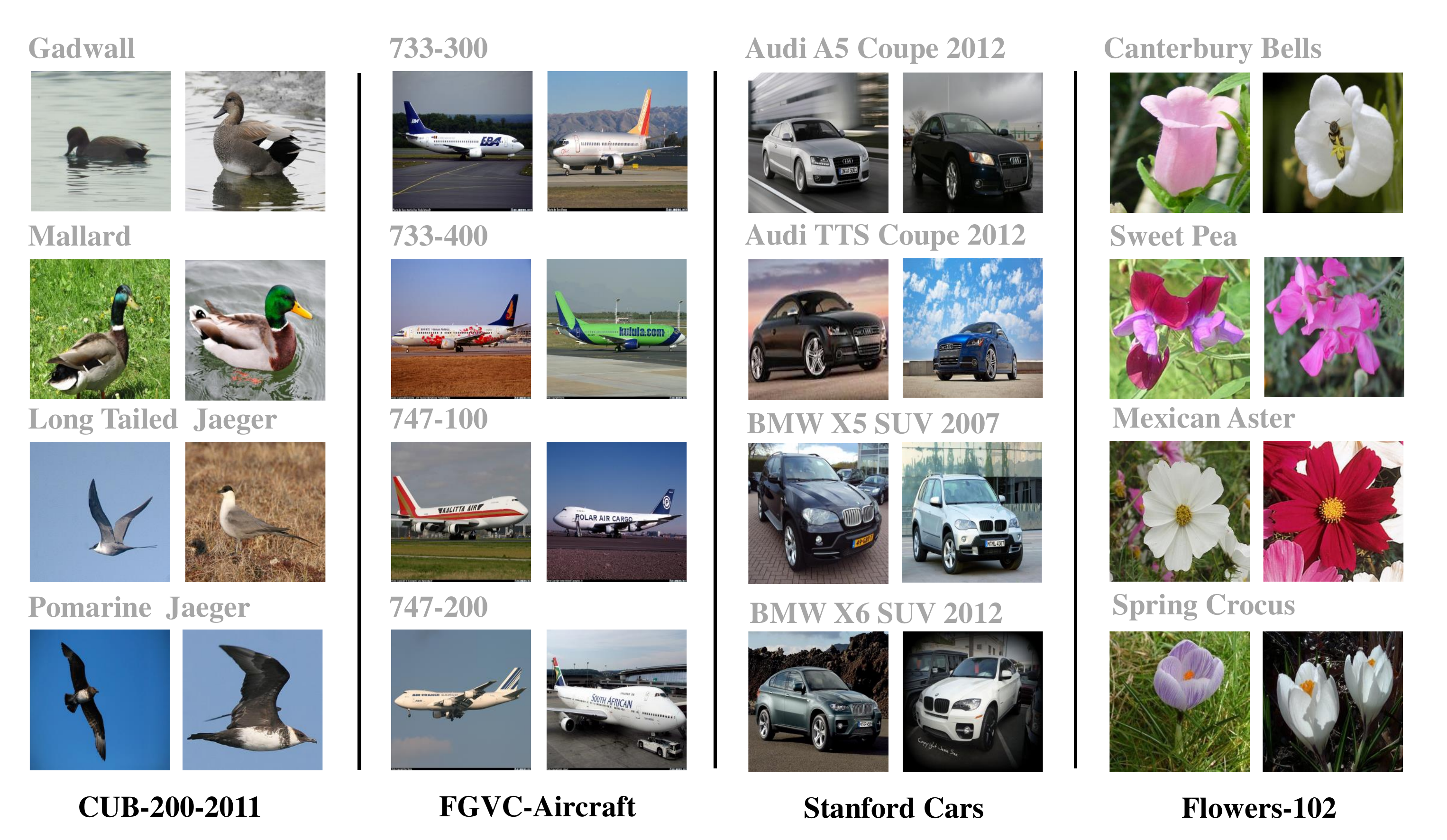}

    \end{center}
    \caption{Sample images from the benchmark datasets: CUB-$200$-$2011$, FGVC-Aircraft, Stanford Cars, and Flowers-$102$}
    \label{fig:dataset}
\end{figure*}

\subsection{Top-down Spatial Attention Loss}\label{section:Cross-layer}


As we all know, convolutional nerual networks (CNNs) hierarchically learn visual patterns from edges and corners to parts and objects. Relative high-level features are obtained through a series of convolutional operations on relative low-level features. In other words, high-level features can be seemed as non-linear combinations of relative low-level features. When the high-level features catch some significative visual patterns (\emph{e.g.}, the head of birds), it is intuitive that we can lead the relative middle-level convolutional layers to search finer discriminative features (\emph{e.g.}, the beak of birds or the crest of birds) in certain semantic regions obtained by high-level features. In reverse, finer discriminative features learned by relative middle-level convolutional layers are benefical to generate more discriminative features at high-level convolutional layers. Both high-level features and relative middle-level features are strengthened in a mutual reinforced way. To do this, we propose the top-down spatial attention loss imposed on features belonging to two different levels, which contains a top-down spatial attention module and a multi-stage channel constrained module.



\subsubsection{The Top-down Spatial Attention Module}

Assuming that the high-level features $F^{h} \in \mathbb{R}^{C^h \times H^h \times W^h}$ catch some significative visual patterns, we can directly use them to constrain the search region of middle-level convolutional filters through the information provided by the high-level features. The output of middle-level convolutional layer can be defined as $F^l \in \mathbb{R}^{C^l \times H^l \times W^l}$, $C^l>=C^h$, $H^l>H^h$, $W^l>W^h$. When the $C^h$ is equal to $C^l$, the top-down spatial attention can be formulated as
\begin{equation}
{F^l}^{\prime} = F^l \otimes Sigmoid(Upsample(F^h)),
\end{equation}
where ${F^l}^{\prime} \in \mathbb{R}^{C^l \times H^l \times W^l}$ denotes the middle-level features with search region constraint and $\otimes$ represents the element-wise multiplication. When the $C^h$ is higher than $C^l$, please see Section~\ref{sec:ablation} for details.
\begin{table}[!t]
  \centering
  \caption{Statistics of datasets.}
    \begin{tabular}{|c|c|c|c|}
    \hline
    Datasets & \#Category & \#Training & \#Test \\
    \hline
    \hline
    CUB-$200$-$2011$ & $200$   & $5994$  & $5794$ \\
    FGVC-Aircraft    & $100$   & $6667$  & $3333$ \\
    Stanford Cars    & $196$   & $8144$  & $8041$ \\
    Flowers-$102$    & $102$   & $2040$  & $6149$ \\
   \hline
    \end{tabular}%
  \label{tab:dataset}%
\end{table}%

\subsubsection{The Multi-stage Channel Constrained Module}
When impose the top-down spatial attention module on the middle-level features, we hope the spatial attention provided by the high-level features can help the middle-level features to find more minute parts, and obtained multi-regional multi-grained features. To accomplish this goal, the feature channels in the middle level and the feature channels in the high-level should one by one aligned. Therefore, we extend the mutual-channel loss~\cite{chang2020mc} to multi-stage, which can make the features channels become category aligned and find multiple discriminative regions, as shown in Figure~\ref{fig:architecture}.a. Thus we define the top-down spatial attention loss ($L_{TDSA}$) as :

\begin{equation}
L_{TDSA} =  L_{{MC}_1}(F^h, y) +  L_{{MC}_2}({F^l}^{\prime},y).
\end{equation}

Then, the total loss function of our network can be formulated as follow:
\begin{equation}
\begin{aligned}
L_{total} & = L_{CE} + {\mu} \times L_{TDSA}. \\
\end{aligned}
\end{equation}



\begin{table*}[!t]
  \centering
  \footnotesize
  \caption{Comparisons of classification accuracies ($\%$) with different loss functions using the VGG$16$ as backbone architecture (trained from scratch). The best and the second best results are respectively marked in red and blue colors.}
  \begin{adjustbox}{width=0.8\linewidth,center}
   \Huge   
    \begin{tabular}{|c|c|c|c|c|c|}
    \hline
    Method     & Base Model             & CUB-$200$-$2011$   & FGVC-Aircraft      & Stanford Cars     & Flowers-$102$ \\
    \hline
    \hline
    CE Loss    & VGG$16$   & $28.53$   & $82.90$    & $76.59$  & $40.90$  \\
Center Loss~\cite{wen2016discriminative}&VGG$16$& $51.38$& $88.26$& $89.27$&$62.53$  \\
A-softmax Loss~\cite{liu2017sphereface}& VGG$16$& $60.79$& $88.15$ & $88.71$& $62.34$  \\
    Focal Loss~\cite{lin2017focal} & VGG$16$& $31.12$  & $80.85$  & $77.02$ & $48.19$ \\
    COCO Loss~\cite{liu2017rethinking} & VGG$16$ & $48.31$& $86.41$ & $67.27$& $63.31$ \\
    LGM Loss~\cite{wan2018rethinking}& VGG$16$& $28.14$ & $87.49$ & $71.27$ & $57.78$  \\
    LMCL Loss~\cite{wang2018cosface}& VGG$16$& $41.11$& $86.17$& $49.57$& $66.43$\\
    ArcFace~\cite{deng2019arcface}& VGG$16$& $36.62$& $82.25$& $79.24$& $48.76$\\
    Circle Loss~\cite{sun2020circle}& VGG$16$& $31.39$& $82.14$& $76.84$& $41.86$\\
    MC-Loss~\cite{chang2020mc} & VGG$16$ & $\textbf{\tblue{65.98}}$ & $\textbf{\tblue{89.20}}$& $\textbf{\tblue{90.85}}$ & $\textbf{\tblue{83.23}}$ \\
    \hline
    \hline
    Ours & VGG$16$& $\textbf{\tred{72.77}}$ & $\textbf{\tred{89.57}}$& $\textbf{\tred{92.29}}$ & $\textbf{\tred{88.74}}$ \\
    \hline
    \end{tabular}%
       \end{adjustbox}
  \label{tab:results_1}%
\end{table*}%

\begin{table*}[!t]
  \centering
  \small
  \caption{Comparisons of classification accuracies ($\%$) with different loss functions using the ResNet$18$ as backbone architecture (trained from scratch). The best and the second best results are respectively marked in red and blue colors.}
    \begin{adjustbox}{width=0.8\linewidth,center}
   \Huge   
    \begin{tabular}{|c|c|c|c|c|c|}
    \hline
    Method     & Base Model             & CUB-$200$-$2011$   & FGVC-Aircraft      & Stanford Cars     & Flowers-$102$ \\
    \hline
    \hline
    CE Loss  &  ResNet$18$   & $45.70$  &  $79.90$   &  $79.12$  & $65.75$  \\
Center Loss~\cite{wen2016discriminative} &   ResNet$18$&  $50.26$  & $83.86$  & $81.84$    & $69.51$  \\
A-softmax Loss~\cite{liu2017sphereface}  &  ResNet$18$    & $49.67$  & $82.42$ &  $82.15$    & $50.56 $ \\
Focal Loss~\cite{lin2017focal}  &  ResNet$18$    & $47.67$ & $80.47$  & $79.75$& $66.87$  \\
COCO Loss~\cite{liu2017rethinking} &  ResNet$18$ & $46.01$  & $80.02$& $72.38$& $66.76$ \\
LGM Loss~\cite{wan2018rethinking} &  ResNet$18$ &  $44.91$&$80.98$&  $74.37$& $66.84$  \\
LMCL Loss~\cite{wang2018cosface} & ResNet$18$& $46.01$&  $78.52$&  $71.17$ &  $67.72$ \\
ArcFace~\cite{deng2019arcface}& ResNet$18$    & $46.67$& $80.33$& $78.32$& $65.68$\\
Circle Loss~\cite{sun2020circle}& ResNet$18$& $47.43$& $79.47$& $78.21$& $66.27$\\
MC-Loss~\cite{chang2020mc}&   ResNet$18$& $\textbf{\tblue{59.41}}$&  $\textbf{\tblue{85.57}}$& $\textbf{\tblue{87.47}}$ &  $\textbf{\tblue{79.54}}$ \\
    \hline
    \hline
Ours &  ResNet$18$ & $\textbf{\tred{69.24}}$ &  $\textbf{\tred{ 86.18}}$ & $\textbf{\tred{ 90.38}} $& $\textbf{\tred{ 85.34}} $\\
    \hline
    \end{tabular}%
           \end{adjustbox}
  \label{tab:results_2}%
\end{table*}%

\begin{figure*}[!t]
  \centering
   \begin{subfigure}[t]{3.2in}
    \centering
    \includegraphics[width=3.2in]{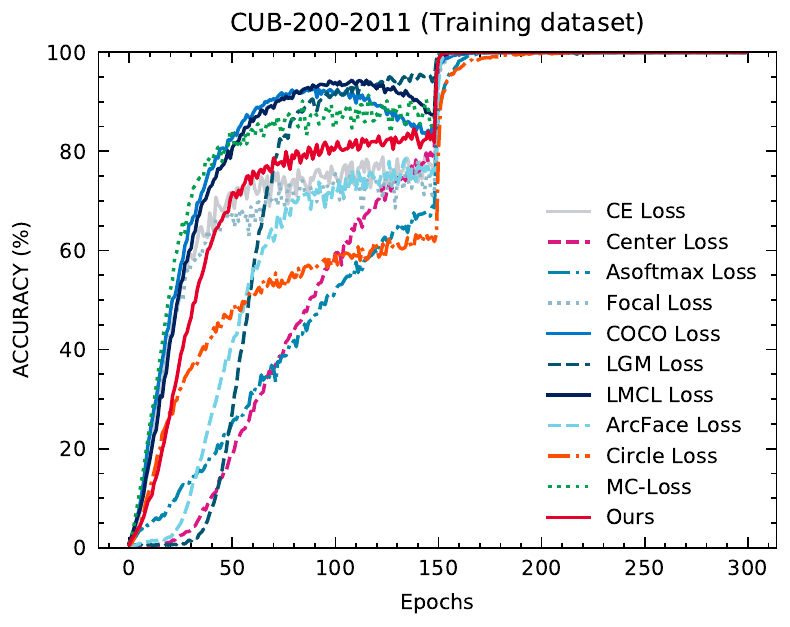}
 \end{subfigure}
  \begin{subfigure}[t]{3.2in}
    \centering
    \includegraphics[width=3.2in]{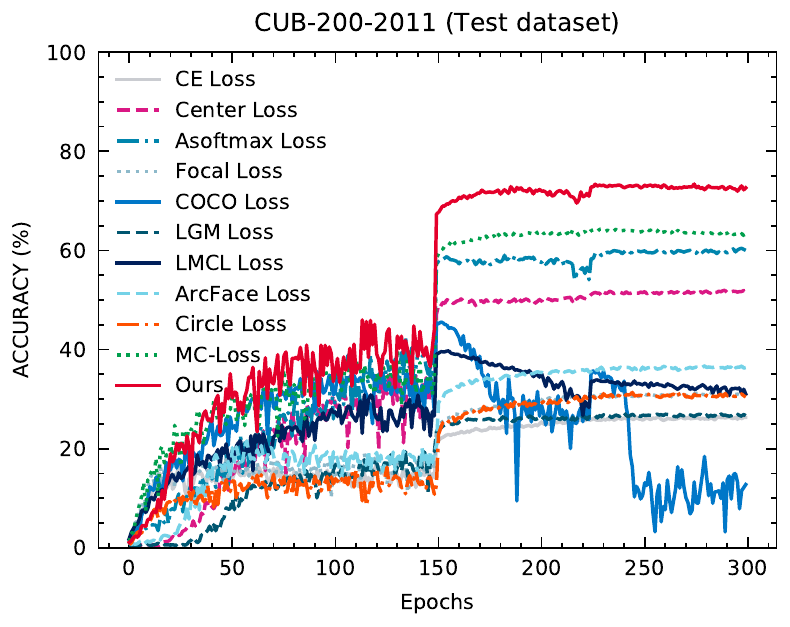}
 \end{subfigure}

  \caption{The accuracies of the proposed method and the other commonly used loss functions on the CUB-$200$-$2011$ dataset using the VGG$16$ as backbone.}
  \label{fig:acc_curve}
\end{figure*}

\section{Experimental Results}

In this section, we firstly introduce the datasets in Section~\ref{sec:dataset} and then present the implementation details in Section~\ref{sec:Details} .  Subsequently, we provided the comprehensive compare results of the proposed methods with other exiting loss functions in Section~\ref{sec:Comparisons}.  As the proposed method contains many modules,   we also provided a comprehensive ablation study in Section~\ref{sec:ablation}.

\subsection{Datasets}\label{sec:dataset}

We evaluated the proposed method on four widely used fine-grained image classification datasets and only used the category level label in our experiment. Details are as follows: (i) CUB-$200$-$2011$ is a bird species dataset and contains $11877$ images belonging to $200$ classes. (ii) FGVC-Aircraft dataset, which contains $10000$ images covering $100$  aircraft models. (iii) Stanford Cars contains $8144$ car image classes by $196$ car models.  (iv) Flowers-$102$  contains $8189$ images belonging to $102$ classes. We follow the standard training/test splits in the original datasets to train and evaluate the proposed method and other exiting loss functions.  A detailed summary of the datasets is provided in Table~\ref{tab:dataset}. Sample images from datasets used are shown in Figure~\ref{fig:dataset}.

\begin{table*}[!t]
  \centering
  \small
  \caption{Ablation study of the proposed method on four fine-grained image classification datasets. (trained from scratch). The best and the second best results are respectively marked in red and blue colors.}
      \begin{adjustbox}{width=0.8\linewidth,center}
   \Huge   
    \begin{tabular}{|c|c|c|c|c|c|}
    \toprule
    Method     & Base Model             & CUB-$200$-$2011$   & FGVC-Aircraft      & Stanford Cars     & Flowers-$102$ \\
    \midrule
    \midrule
    CE Loss & VGG$16$ & $28.53$ & $82.90$  & $76.59$ &$ 40.90$ \\
    MC-Loss & VGG$16$ & $65.98$ & $\textbf{\tblue{89.20}}$  & $\textbf{\tblue{90.85}}$ & $83.23$ \\
    Ours w/o attention & VGG$16$ &$ \textbf{\tblue{69.95}}$ & $88.89$ &$ 90.61$ &$ \textbf{\tblue{85.81}}$ \\
    Ours  & VGG$16$ & $\textbf{\tred{72.77}} $& $\textbf{\tred{89.57}}$ &$ \textbf{\tred{92.29}}$ &$ \textbf{\tred{88.74}}$ \\
    \bottomrule
    \end{tabular}%
               \end{adjustbox}
  \label{tab:ablation_1}%
\end{table*}%

\begin{table*}[!t]
  \centering
  \small
  \caption{Influence of feature channel number on four fine-grained image classification datasets using the VGG$16$ as backbone architecture. $\xi =i$ means each category has $i \times 3$ feature channels in the middle-level.}
  \begin{adjustbox}{width=0.8\linewidth,center}
   \Huge   
    \begin{tabular}{|c|c|c|c|c|c|}
    \toprule
    Method     & Base Model             & CUB-$200$-$2011$   & FGVC-Aircraft      & Stanford Cars     & Flowers-$102$ \\
    \midrule
    \midrule
    Ours with $\xi =1$ & VGG$16$ & $71.06$        & $88.82$        & $91.68$        & $87.14$ \\
    Ours with $\xi =2$ & VGG$16$ & $72.12$        & $\textbf{\tred{89.57}}$ & $\textbf{\tred{92.29}}$ & $87.81$ \\
    Ours with $\xi =3$ & VGG$16$ & $\textbf{\tred{72.77}}$ & $\textbf{\tblue{89.21}}$& $\textbf{\tblue{92.08}}$& $\textbf{\tred{88.74}}$ \\
    Ours with $\xi =4$ & VGG$16$ & $\textbf{\tblue{72.76}}$& $89.00$        & $91.37$        & $\textbf{\tblue{88.26}}$ \\
    Ours with $\xi =5$ & VGG$16$ & $72.58$        & $88.76$        & $91.94$        & $87.60$ \\

    \bottomrule
    \end{tabular}%
  \end{adjustbox}
  \label{tab:ablation_2}%
\end{table*}%

\begin{table*}[!t]
  \centering
  \small
  \caption{Influence of feature channel number on four fine-grained image classification datasets using the ResNet$18$ as backbone architecture. $\xi =i$ means each category has $i \times 3$ feature channels in the middle-level.}
  \begin{adjustbox}{width=0.8\linewidth,center}
   \Huge   
    \begin{tabular}{|c|c|c|c|c|c|}
    \toprule
    Method     & Base Model             & CUB-$200$-$2011$   & FGVC-Aircraft      & Stanford Cars     & Flowers-$102$ \\
    \midrule
    \midrule
    Ours with $\xi =1$ & ResNet$18$ & $65.25$        & $84.65$        & $88.37$        & $82.86$ \\
    Ours with $\xi =2$ & ResNet$18$ & $68.72$        & $\textbf{\tred{86.18}}$ & $\textbf{\tred{90.38}}$ & $84.34$ \\
    Ours with $\xi =3$ & ResNet$18$ & $\textbf{\tred{69.24}}$& $\textbf{\tblue{86.09}}$& $\textbf{\tblue{89.70}}$& $\textbf{\tred{85.34}}$ \\
    Ours with $\xi =4$ & ResNet$18$ & $\textbf{\tblue{69.12}}$ & $85.55$        & $89.08$        & $\textbf{\tblue{85.32}}$ \\
    Ours with $\xi =5$ & ResNet$18$ & $68.72$        & $85.87$        & $89.36$        & $83.78$ \\

    \bottomrule
    \end{tabular}%
  \end{adjustbox}
  \label{tab:ablation_3}%
\end{table*}%

\subsection{Implementation Details}\label{sec:Details}

For fair comparisons, we adapted the ResNet$18$ and VGG$16$ model as the backbone model and resized each input image to $224\times224$ throughout the experiments.  We use Stochastic Gradient Descent optimizer and batch normalization as the regularizer. We train the model from scratch for $300$ epochs, and the value of weight decay is kept as $5e-4$. The model's learning rate is initially set as $0.1$ and multiplied by $0.1$ at $150^{th}$ and $225^{th}$ epoch, successively. Furthermore, we set the hyper-parameters of the proposed method as $\mu = 1.5$ and $\lambda = 10$. Especially, follow the suggestions of the MC-Loss~\cite{chang2020mc}: in the high-level, we use $3$ channels to represent one class; in the middle-level, the channel number is higher than the high-level because there are many small parts in a region, see Section~\ref{sec:ablation} for details.

\begin{table}[!t]
  \centering
  \small
  \caption{Comparisons of classification accuracies (\%) with different upsample methods using the VGG$16$ as backbone architecture. $\xi =i$ means each category has $i \times 3$ feature channels in the middle-level.}
  \begin{adjustbox}{width=0.8\linewidth,center}
   \Huge   
    \begin{tabular}{|c|c|c|c|}
    \toprule
    Method & Base Model & Upsample & Acc. \\
    \midrule
    \midrule
    Ours with $\xi =1$ & VGG$16$ & nearest & $73.33 $\\
    Ours with $\xi =1$ & VGG$16$ & bicubic & $72.82 $\\
    Ours with $\xi =1$ & VGG$16$ & bilinear &$ 73.03$ \\
    \bottomrule
    \end{tabular}%
  \end{adjustbox}
  \label{tab:ablation_4}%
\end{table}%

\begin{table}[!t]
  \centering
  \small
  \caption{Comparisons of classification accuracies (\%) with different upsample methods using the ResNet$18$ as backbone architecture. $\xi =i$ means each category has $i \times 3$ feature channels in the middle-level.}
  \begin{adjustbox}{width=0.8\linewidth,center}
   \Huge   
    \begin{tabular}{|c|c|c|c|}
    \toprule
    Method & Base Model & Upsample & Acc. \\
    \midrule
    \midrule
    Ours with $\xi =1$ & ResNet$18$ & nearest  & $65.66$ \\
    Ours with $\xi =1$ & ResNet$18$ & bicubic  & $65.50$ \\
    Ours with $\xi =1$ & ResNet$18$ & bilinear & $65.33$ \\
    \bottomrule
    \end{tabular}%
  \end{adjustbox}
  \label{tab:ablation_5}%
\end{table}%

\subsection{Comparisons With State-of-the-Art Methods}\label{sec:Comparisons}
Table~\ref{tab:results_1} and~\ref{tab:results_2}  shows the comparison results between the proposed loss functions and other exiting loss function on the four widely used fine-grained image classification dataset.  From Table~\ref{tab:results_1}, we can observe that when using VGG16 as the backbone, we obtained the best results of 72.77\%, 89.57\%, 92.29\%, and 88.74\% on CUB-$200$-$2011$, FGVC-Aircraft, Stanford Cars, and Flowers-102 datasets, respectively.  Similar results can also find in Table~\ref{tab:results_2}. We can see that the proposed method still obtained the best performance on four fine-grained image classification datasets using the ResNet$18$ as the feature extractor. In summary, the proposed method defeated all the compared methods on the four widely used fine-grained image classification datasets for both VGG16 and ResNet18 backbone. Meanwhile, Figure~\ref{fig:acc_curve} illustrated the proposed method and the other commonly used loss functions' accuracies curves on the CUB-$200$-$2011$ dataset. From Figure~\ref{fig:acc_curve}, the proposed method improved the optimization characteristics and obtained consistent gains in performance.

\begin{figure*}[!t]
    \begin{center}
   \includegraphics[width=0.95\linewidth]{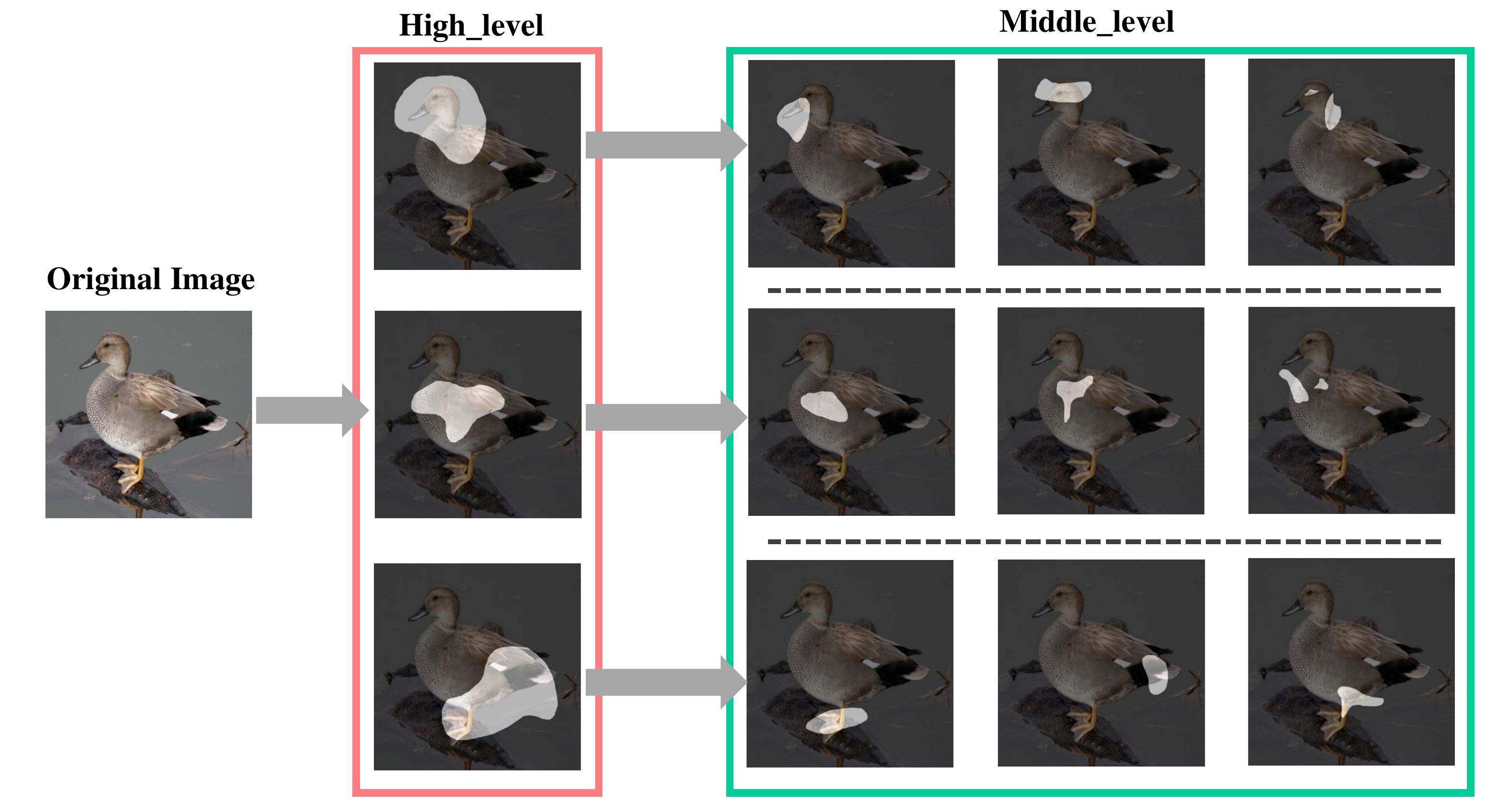}

    \end{center}
    \caption{Channel visualizations ($\xi=3$). $\xi =i$ means each category has $i \times 3$ feature channels in the middle-level. The first column represents the original image. the second column is the channel visualization of the high-level channels belonging to one class; the third to the fifth column is the channel visualization of the middle-level channels belonging to the same class. All high-level channels are in the red box and all middle-level features are in the green box.}
    \label{fig:visualization_MC}
\end{figure*}

\subsection{Ablation Study}\label{sec:ablation}

To further demonstrated the effectiveness of the proposed method, we conducted an ablation study using VGG16 as the backbone to justify the contribution of each module of the proposed method,  as shown in Table~\ref{tab:ablation_1}. (i) Compared with MC-Loss, ``Ours w/o attention'' obtained higher performance on the CUB-200-2011  and Flowers-102 datasets, slightly worse performance on the FGVC-Aircraft and Stanford Cars datasets.  This indicates that the multi-stage channel constrained module can help the model improve the performance on the flexible object (\emph{e.g.}, bird and flower) but decrease the performance on the rigid object (\emph{e.g.}, car and aircraft). (ii) The proposed ``Ours'' obtained higher accuracy than ``Ours w/o attention'' on all datasets, which demonstrates that the top-down spatial attention module can help the multi-stage channel constrained module capture more specific local regions and assist the model work well with the rigid objects. Hence, better performance can be obtained. Furthermore, the top-down spatial attention module cannot work if there is no multi-stage channel constrained module to make the high-level and middle-level feature channels become category aligned.

\subsubsection{Influence of the $\xi$}

To evaluate the influence of $\xi$ on the accuracy, we vary $\xi$ from $1$ to $5$ uniformly. $\xi =i$ means each category has $i \times 3$ feature channels in the middle-level. Therefore, if the $\xi$ is higher than one, the high-level channels will be less than the middle-level channels; in this situation, we can repeat each high-level channel $\xi$ times. 
From Tabel~\ref{tab:ablation_2}, we can see that the proposed method obtained the best results on the CUB-$200$-$2011$ and Flower-$102$ datasets when the $\xi = 3$, and the performance is dropped when we increase the value of $\xi$. Those phenomena indicate that when applying the proposed method to recognize the flexible object, a higher $\xi$ is better, but a  too high $\xi$ may increase the difficulty of model optimization due to there don not exiting enough discriminative parts to mine. Meanwhile, we can observe that the proposed method obtained the best results on the FGVC-Aircraft and Stanford Cars datasets when the $\xi$ =$2$, and the performance is also dropped when we increase the value of $\xi$.  Those phenomena indicate that when applying the proposed method to recognize the rigid object, a lower $\xi$ is better, but a too low or too high $\xi$ may hurt the proposed method's performance.  We can from Table~\ref{tab:ablation_3} find a similar phenomenon.

\subsubsection{Influence of the Upsampling Method}

In the top-down spatial attention mechanism, we need to upsample the attention map obtained by the high-level channels, then using it to supervise the middle-level channels.  Therefore, in this section, we investigate the influence of the upsample methods (\emph{e.g.}, nearest, bicubic, and bilinear).  Table~\ref{tab:ablation_4} and~\ref{tab:ablation_5} show that the different upsample methods have a minute influence on the results, no matter what network we use as the backbone.

\subsubsection{Visualization}
To illustrate the advantages of the proposed method intuitively, we visualize the high-level and middle-level channels.  Note that the proposed method will make the high-level and middle-level channels become category aligned. Therefore, we do not need to use the Grad-CAM~\cite{selvaraju2017grad} technique to visualize the channels; we can directly visualize the channels belonging to each class. As shown in Figure~\ref{fig:visualization_MC}, the first column is the original image; the second column is the channel visualization of the high-level channels belonging to one class; the third to the fifth column is the channel visualization of the middle-level channels belonging to the same class. Especially in each row,  the second column channel visualization supervises the third to the fifth channels. We can see that: (i) the three high-level channels learned different and discriminative regions; (ii) under the supervise of the high-level channels, the middle-level channels learned more subtle parts and different from each other; (iii) with the help of the proposed method, the model learned multi-regional multi-grained features, which can help the model to improve the performance. 

\section{Discussions}
In this paper, we argue that learning multi-regional multi-grained features is the key for FGVC. Follow this idea,  we proposed a new loss function, namely the TDSA Loss, to address this problem. The proposed framework obtained the best results on four widely used fine-grained image classification datasets.  

This paper~\emph{only} focuses on the middle-level and high-level features channels because they have clear semantic information. The high-level features contain global information, and the middle-level features contain local information. We can also see this from Figure~\ref{fig:visualization_MC}, and the results also demonstrate the choice is reasonable.  However, there exists an open problem: what will happen when we focus on more level channels, rather than only for the middle-level and the high-level? The answer depends on the change of the receptive field. Taking the VGG16 as an example, which contains five convolutional blocks. In our experiments, we treat the fourth convolutional block's output as the middle-level features and the fifth convolutional block's output as the high-level features. It is worth noting that each convolutional block has the same receptive field. Therefore, we can ignore the features between the middle-level and high-level. Meanwhile, from Figure~\ref{fig:visualization_MC}, we can see that the regions of middle-level features are very small. Thus, we can ignore the feature before the fourth convolutional block. The low-level features contain some texture and shape features, which are useless for FGVC. Those phenomenons still exist in other networks, ~\emph{e.g.}, ResNet$18$.

\section{Conclusion}
In this paper, we show that the key to fine-grained image classification is to explore the multi-regional multi-grained features. The proposed method can effectively drive the high-level and the middle-level feature channels to be more discriminative and focusing on multi-regional multi-grained features, without the need for fine-grained bounding-box/part annotations. We also show that different networks can easily integrate the proposed method to improve performance. Experiments on all four fine-grained image classification datasets have demonstrated the superiority of the proposed method. In the future, we will investigate to apply the proposed method to other tasks, such as fine-grained image retrieval.


%



\ifCLASSOPTIONcaptionsoff
  \newpage
\fi



%

\bibliographystyle{IEEEbib}
\bibliography{main}

\end{document}